\documentclass[letterpaper, 10 pt, conference]{ieeeconf}

\IEEEoverridecommandlockouts                              %

 \usepackage{graphicx}
 \usepackage{xcolor}
 \usepackage{censor}
 \usepackage{url}
 \usepackage{algorithm}
\usepackage{algpseudocode}
\usepackage{float}
\usepackage{amsmath} %
\usepackage{amssymb}  %
\usepackage{siunitx}
\usepackage{subcaption} %

\title{\LARGE \bf AeRove: A Compact Bimodal Aerial-Terrestrial Drone with \\Rapid Bistable Reconfiguration for Close-Range Pipeline Inspection}%

\author {Caleb Polillio$^{1}$ and Petras Swissler$^{1}$%
\thanks{*This work was supported by The National Science Foundation and The New Jersey Institute of Technology}%
\thanks{$^{1}$Caleb Polillio and Petras Swissler are with the Department of Mechanical and Industrial Engineering, New Jersey Institute of Technology, Newark, NJ 07102
        {\tt\small petras.swissler@njit.edu}}%
}

\begin{document}

\maketitle
\thispagestyle{empty}
\pagestyle{empty}

\begin{abstract}

Close-proximity pipeline inspection is challenging for standard drones due to high hovering power consumption, airflow sensitivity, and propeller wash interference with gas sensing. To address this, we present AeRove, a compact bimodal aerial-terrestrial robot. AeRove uses its propeller guards as wheels to roll along pipes and employs a spring-loaded bistable mechanism to reconfigure between ground and flight modes in 200 ms without continuous actuator power to maintain either state. The converging propeller-guard geometry also improves measured thrust efficiency.
By reserving flight for obstacle hopping and using ground rolling for continuous traversal, current draw is reduced 14$\times$ compared to continuous flight (\qty{0.7}{\A} vs. \qty{10}{\A}), extending estimated travel distance from \qty{144}{m} to \qty{2057}{m}. Autonomous trials on a \qty{51}{cm} diameter steel pipe demonstrated navigation on straight and curved sections, obstacle jumping, and leak detection of simulated inspection markers. Separate CO\textsubscript{2} sensing experiments evaluated gas-detection performance under perched and aerial operating conditions. Perched inspection produced a substantially larger concentration response than hovering under the tested conditions. During flight, an underbody propeller-intake configuration produced a faster and stronger gas-detection response than an extended probe by leveraging propeller-induced airflow. Code and designs are released under the CC-BY license.

\end{abstract}

\section{Introduction}

Multi-rotor drones are an invaluable tool for monitoring and inspection, since they enable remote observation of otherwise difficult-to-reach locations. For example, the New York City Fire Department has used drones since 2017 for emergency reconnaissance %
\cite{leo2019fdnydrones}. In agriculture, using drone-based data collection and crop monitoring enables farmers to better understand their fields and increase crop yields \cite{nazarov2023drones}. In industrial settings, drones perform inspection in hard-to-reach locations such as high-up locations and confined spaces, where human inspection would be unsafe \cite{pinninti2023using, tripicchio2018confined}.

\begin{figure}[]%
    \centering
    \includegraphics[width=\linewidth, trim=0 6cm 1cm 5cm, clip]{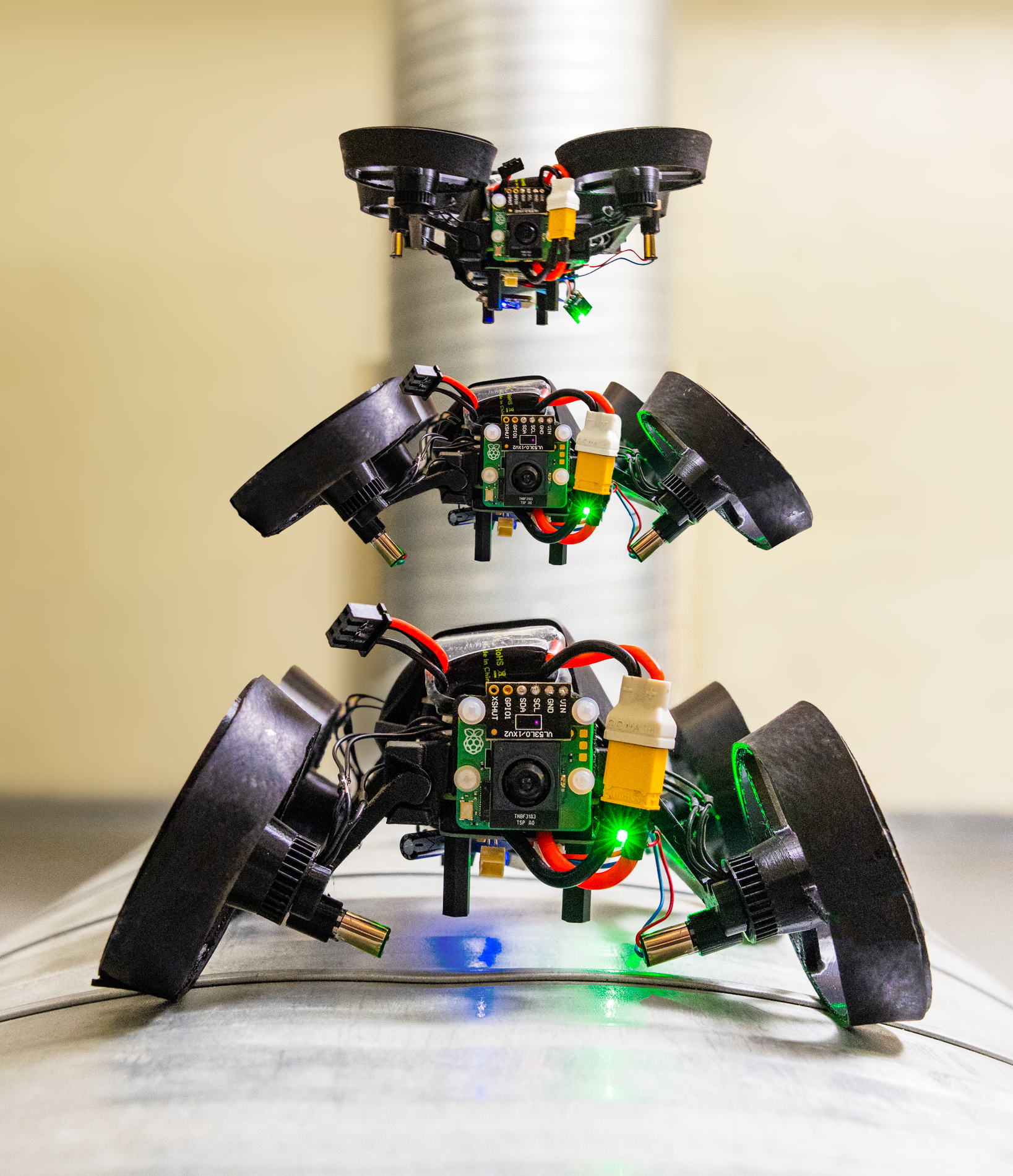}
    \caption{Composite front view of AeRove transforming from flight mode to rover mode as it lands on a ventilation duct.}
    \label{fig:hero}
\end{figure}

Flight control of drones is well-studied and understood  \cite{idrissi2022review, peksa2024review} with many off-the-shelf flight controllers (e.g., \cite{micoair743v2_aio_45a}). Unfortunately, station keeping when experiencing sudden changes in air movement remains a challenge. Outside of yet-uncommon drones that proactively sense air movement \cite{kent2026asymmetric, simon2023flowdrone} controllers only resist errors that are already occurring, making this challenge particularly pronounced in small drones, which generally have lower thrust capacity and less inertia. Poor station-keeping is problematic during close-proximity inspection, where precision is required for reliable sensing and where proximity to the inspected structure increases collision risk.
While larger and advanced drones can station keep with greater accuracy, this size inherently limits the spaces they can enter to inspect. Perching drones have an inherent ability to station-keep once they land, but since they are stationary when landed, the range of sensing is likewise inherently limited \cite{wuest2024agile, meng2022aerial, li2025treecreeper}. %

\begin{figure*}[]
    \centering
    \includegraphics[width=0.8\linewidth]{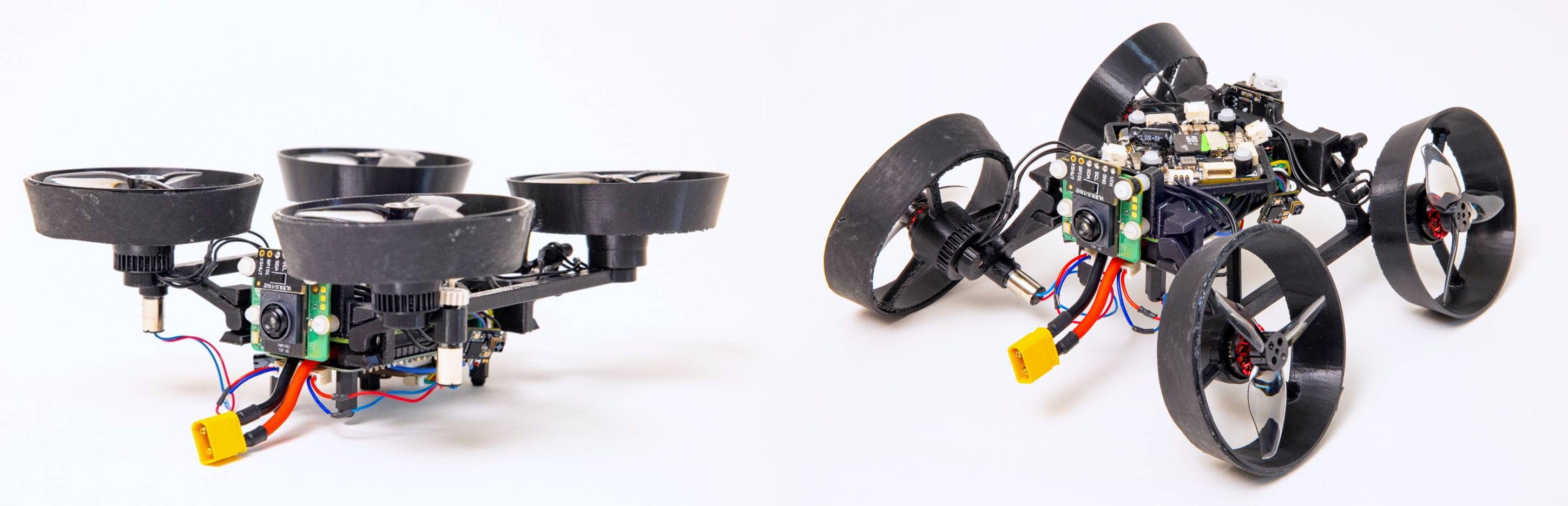}
    \caption{AeRove shown without its battery in (Left) flight mode and (Right) terrestrial mode.}
    \label{fig:isoview}
\end{figure*}

A type of drone that has emerged in recent years is the bimodal drone capable both of terrestrial and aerial motion \cite{sihite2023m4,flyingstar2,tang2025duawlfin}.%
We propose that this type of drone would be very useful for close-proximity inspection tasks, since such a drone could fly into position, land, and perform inspections. Specifically, we believe that the use case of pipe inspection would be an ideal application since a small bimodal drone would be able to perform close-proximity inspection (e.g., crack monitoring or chemical sensing) by rolling along the pipe surface, then using its flying mode to ``jump'' over valves or otherwise difficult-to-traverse features. This approach also avoids potential measurement issues associated with rotor airflow blowing away the very chemical signatures that are being sensed. %
Examples of drones that use multi-modal motion for inspection include FalconScan a heterogeneous flyer-crawler team \cite{abdellatif2024falconscan}, circumferential (rather than longitudinal) traversing of a pipe using thrust adhesion \cite{watson2025fly}, and a drone that wraps tentacles around the pipe being inspected \cite{garcia2021soft}.

\section{Design}

In this paper, we present AeRove, a lightweight hybrid aerial-terrestrial robotic platform designed for close-range inspection in environments where continuous flight is unnecessary or inefficient. Fig. \ref{fig:isoview} shows it both in flying mode and in terrestrial mode. %
A bistable transition mechanism rotates the appendages between ground and flight configurations while mechanically holding either state without continuous actuator input. This multifunctional use of the its propeller guards as wheels reduces the need for separate locomotion hardware, limiting additional mass while preserving a compact geometry and enabling rapid switching between ground and flight modes in confined inspection environments.

Relevant design files and software have been released open-source under the CC-BY license \cite{aerove2026repository}. The overall cost of AeRove is \$468 when purchasing individual components. This compares favorably to the commonly-used, flying-only Crazyflie 2.1+ platform, which costs \$240 with with the advantage of bulk purchasing \cite{bitcraze_crazyflie21plus_2026}.

\subsection{Mechanical Design}

AeRove's primary structural components were FDM 3D printed using PETG, enabling rapid design iteration while providing the toughness needed for repeated ground contact and flight testing. The vehicle also uses a modular architecture, allowing the wheel, transition, and propulsion assemblies to be replaced or modified independently.

\subsubsection{Drive}

Terrestrial locomotion is provided by two \qty{6}{mm} planetary gear motors with 136:1 reduction, each driving one front wheel through an additional 2:1 gear stage. The wheels have a diameter of \qty{5.9}{\centi\meter}, maintaining sufficient clearance from the propellers to avoid interference. Each wheel is supported by a ball bearing to reduce rotational friction in both the driven and passive wheel assemblies. Latex treads were wrapped and bonded to the driven wheels to provide traction, while the passive wheels are left untreaded to facilitate a differential drive configuration. The two-wheel-drive configuration enables zero-radius turns without a dedicated steering mechanism. Compliance during ground locomotion is provided by the spring-loaded transition assembly described in the following subsection.

\begin{figure}[t]
    \centering

    \includegraphics[width=\linewidth]{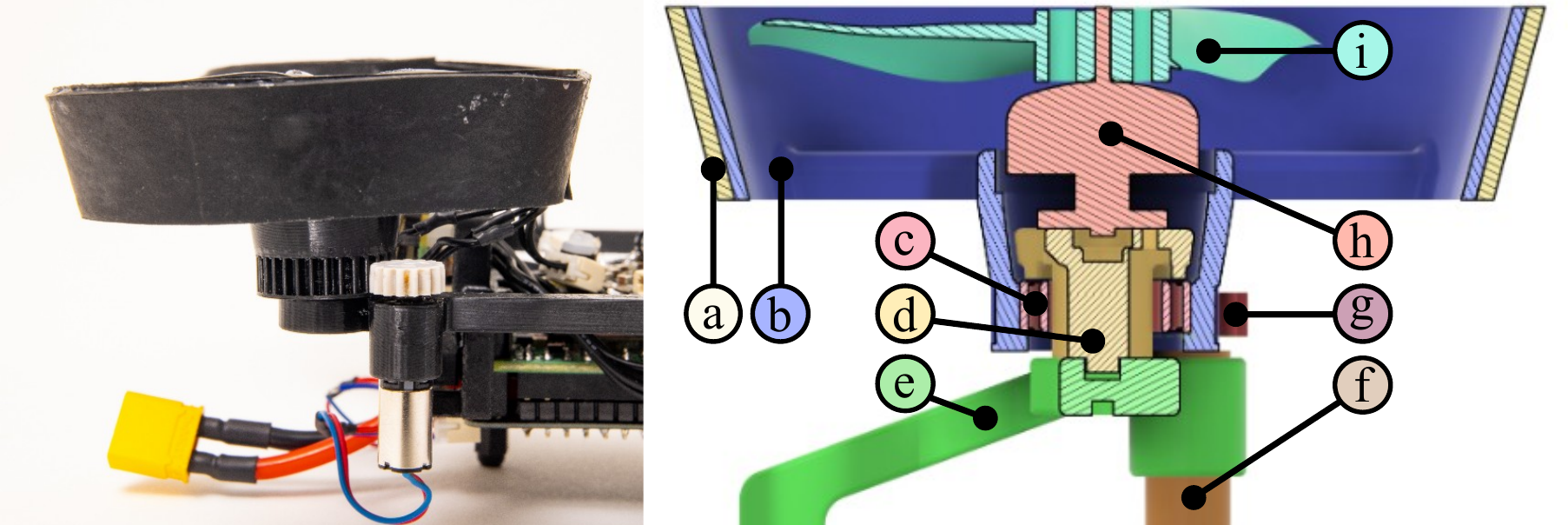}

    \caption{(Left) Photo, terrestrial drive wheel and motor in the flying configuration. %
    (Right) CAD section view showing the (a) Tire; (b) Wheel; (c) Bearing; (d) Motor mount; (e) Arm to the main body; (f) Terrestrial drive DC motor; (g) Drive gear; (h) Flying brushless DC motor; (i) Propeller.}
    \label{fig:side_by_side}
\end{figure}

\subsubsection{Transition}

AeRove transitions between locomotion modes using a spring-loaded bistable mechanism actuated by a \qty{2.9}{\gram} linear servo \cite{li2021soft}. A rigid linkage couples the servo motion to the rotating appendages, while a spring provides the restoring force required to maintain two mechanically stable configurations corresponding to ground and flight modes. Both stable configurations are physically bounded by mechanical stops, which define the two retained endpoint states of the mechanism. During a ground-to-flight transition, the servo drives the linkage through an energy barrier as the spring is extended. Once the mechanism passes the unstable intermediate configuration, the spring assists the remaining motion and holds the appendages in the flight state without continuous actuator input.  The reverse process occurs during the flight-to-ground transition. In the ground configuration, the same spring-loaded linkage also provides suspension during terrestrial locomotion. Because the wheels are mounted with a 15$^\circ$ camber, the appendages require only 75$^\circ$ of rotation between the flight and ground configurations to be parallel with the ground, rather than a full 90$^\circ$. 

\begin{figure*}[]
    \centering
    \includegraphics[width=0.9\linewidth]{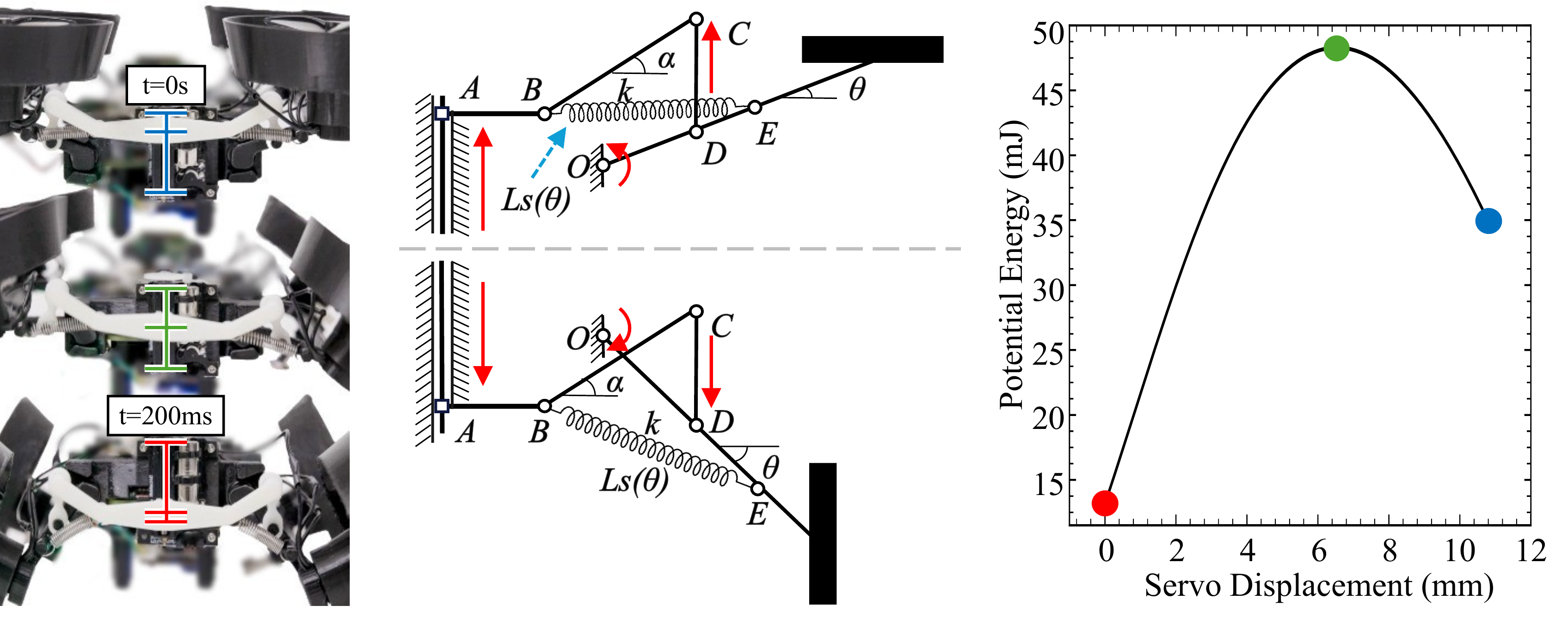}
    \caption{(Left) Physical transition mechanism,  (Center) mechanical model in drive and flight mode (see Table \ref{tab:mechanism_parameters}), (Right) graph of spring energy vs displacement of the servo. Colors in left photos and right graph correspond.}
    \label{fig:design}
\end{figure*}

The bistable linkage was modeled using planar rigid-body kinematics to characterize the relationship between actuator displacement, spring extension, and elastic potential energy. The rigid-body constraint imposed by the connecting link is:

\begin{equation}
\begin{split}
&\left[ -s_x+a+b\cos\alpha-r_D\cos\theta \right]^2 \\
&\quad + \left[ -s_y+q+b\sin\alpha-r_D\sin\theta \right]^2 = l^2.
\end{split}
\end{equation}

Servo displacement \(q\) is referenced from the ground-mode mechanical stop, such that \(q=0\) corresponds to the ground configuration. Solving for actuator displacement gives

\begin{equation}
\begin{split}
q(\theta) &= s_y - b\sin\alpha + r_D\sin\theta \\
          &\quad + \sqrt{l^2 - \left[ -s_x+a+b\cos\alpha-r_D\cos\theta \right]^2}
\end{split}
\end{equation}

The corresponding spring length is

\begin{equation}
\begin{split}
L_s(\theta) = \\
    \sqrt{ \left( r_E\cos\theta+s_x-a \right)^2 + \left( r_E\sin\theta+s_y-q(\theta) \right)^2 }.
\end{split}
\end{equation}

and the total elastic potential energy of the mirrored mechanism is:

\begin{equation}
U_{\mathrm{tot}}(\theta)
=
k
\left[
L_s(\theta)-L_0
\right]^2
\end{equation}

evaluated over the operating range $\theta_{\min} \leq \theta \leq \theta_{\max}$

The resulting potential-energy profile is shown in Fig.~\ref{fig:design}, and the geometric and spring parameters used in the model are summarized in Table~\ref{tab:mechanism_parameters}.

\begin{table}[!t]
\centering
\caption{Bistable Mechanism Parameters}
\label{tab:mechanism_parameters}
\begin{tabular}{clc}
\hline
\textbf{Symbol} & \textbf{Parameter} & \textbf{Value} \\
\hline
$a$ & $AB$ length & 16.00 mm \\
$b$ & $BC$ length & 12.53 mm \\
$l$ & $CD$ length & 10.00 mm \\
$r_D$ & $OD$ length & 9.417 mm \\
$r_E$ & $OE$ length & 15.832 mm \\
$\alpha$ & Crossbar angle & $26.9^\circ$ \\
$s_x$ & Horizontal $A$-to-$O$ offset & 20.5 mm \\
$s_y$ & Vertical $A$-to-$O$ offset & 3.2 mm \\
$L_0$ & Spring free length & 12.0 mm \\
$k$ & Spring constant & 0.628 N/mm \\
$\theta_{\min}$ & Lower rocker angle & $-55^\circ$ \\
$\theta_{\max}$ & Upper rocker angle & $20^\circ$ \\
\hline
\end{tabular}
\end{table}

The mechanism geometry was selected to account for the different loading conditions associated with each locomotion state. In ground mode, wheel-ground contact transmits external loads into the wheel assembly and suspension linkage, so the mechanism was designed with a larger elastic-energy barrier to reduce the risk of unintended reconfiguration during terrestrial operation. In flight, propeller thrust acts in a direction that helps maintain the flight configuration, allowing a lower energy barrier while still providing stable retention. These external loads are not included directly in the spring-energy model; instead, they informed the asymmetric geometry evaluated by the model.

Ground reconfiguration can be hindered when rough or uneven terrain obstructs the motion of the transition appendages. Because AeRove can complete the transition rapidly, the vehicle can begin reconfiguring to ground mode before touchdown and land directly on its wheels \cite{mandralis2025atmo}. This removes the likelihood that terrain contact will interfere with the transition. The spring-loaded linkage also introduces compliance during touchdown, helping absorb impact loads transmitted through the wheel assemblies.

\subsubsection{Flight}

Aerial locomotion is provided by a MicoAir 743v2 AIO flight controller, Happymodel 11000 KV brushless motors, and Gemfan 2023-3 propellers in a \qty{12.8}{\centi\meter} H-frame configuration. A MicoAir MTF-02P optical-flow and rangefinding module provides local motion and altitude information for stable flight without GPS. The propulsion system is capable of a maximum thrust-to-weight ratio of approximately 2.7:1; however, normal operation is limited to approximately 2:1 to reduce unnecessary power consumption while maintaining sufficient control authority.

The propeller guards converge at an angle of 15° from top to bottom, narrowing beneath the propeller disk based on the hypothesis that, by the continuity equation, this reduced flow area increases the downstream air velocity. Bernoulli's principle then relates this increase in velocity to a decrease in static pressure below the propeller. Since thrust is produced by the pressure difference across the propeller disk, the converging guard geometry can increase this pressure differential relative to a non-converging configuration and improve thrust efficiency. This behavior is consistent with the aerodynamic advantages commonly associated with ducted and shrouded propeller systems.  
Four convergence angles were experimentally evaluated to determine their effect on thrust efficiency: 0$^\circ$, 10$^\circ$, 15$^\circ$, and 30$^\circ$. 

\begin{figure}[]%
    \centering
    \includegraphics[width=0.85\linewidth]{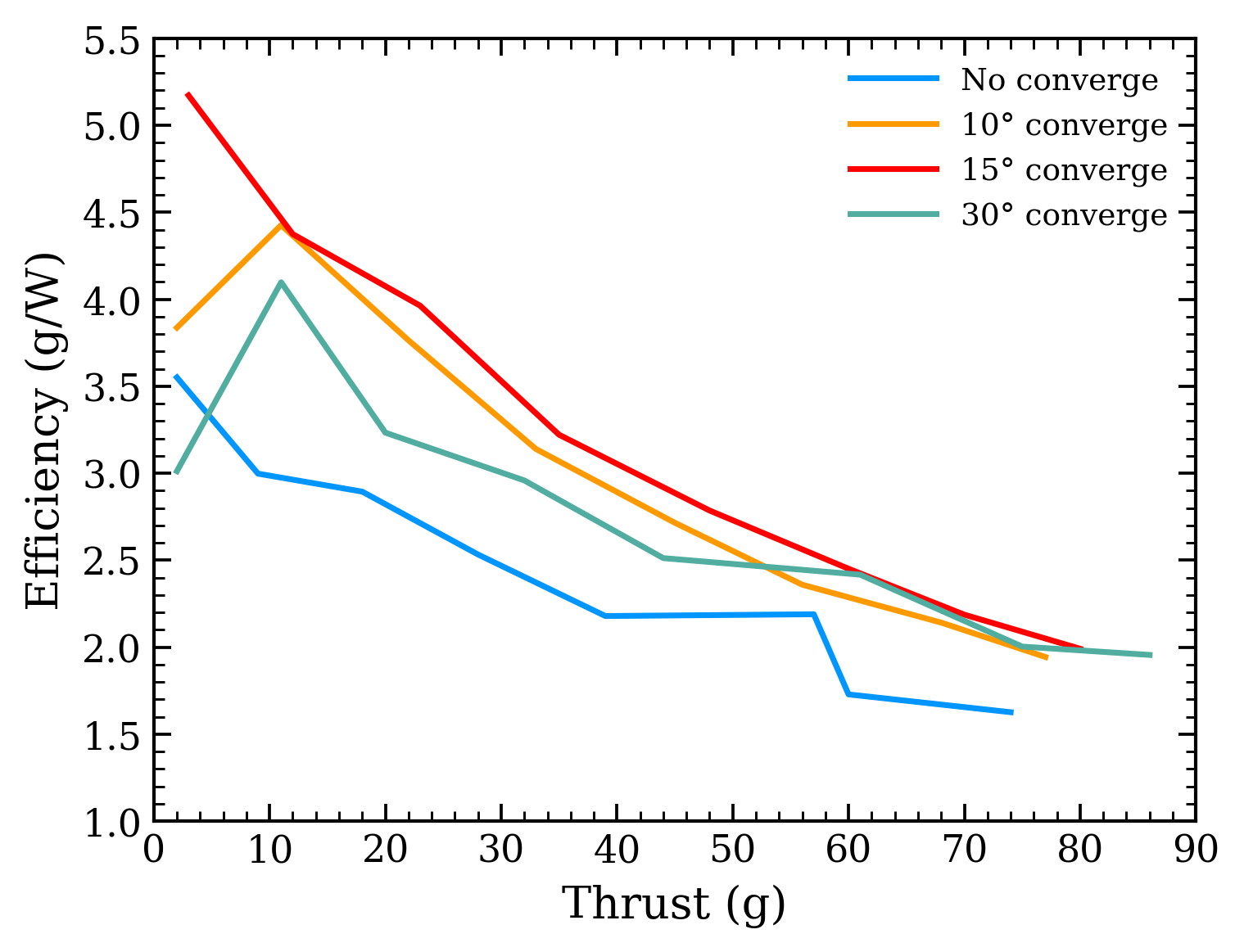}
    \caption{Propeller-guard efficiency as a function of convergence angle.}
    \label{fig:propguardeff}
\end{figure}

As shown in Fig.~\ref{fig:propguardeff}, the 15$^\circ$ configuration produced the highest measured efficiency across the operating range. At a hover thrust setting (approximately \qty{43}{\gram}), the 15$^\circ$ guard improved thrust efficiency by approximately 35\% relative to the 0$^\circ$ configuration. At the 2:1 thrust-to-weight operating point, the improvement was approximately 26\%. Efficiency decreased at $30^\circ$, indicating that excessive constriction reduced performance, likely due to increased flow losses or separation. Among the tested geometries, $15^\circ$ provided the best measured compromise between acceleration of the downstream flow and restriction of the propeller stream.

\subsection{Electrical Design}

\begin{figure}[]%
    \centering
    \includegraphics[width=\linewidth]{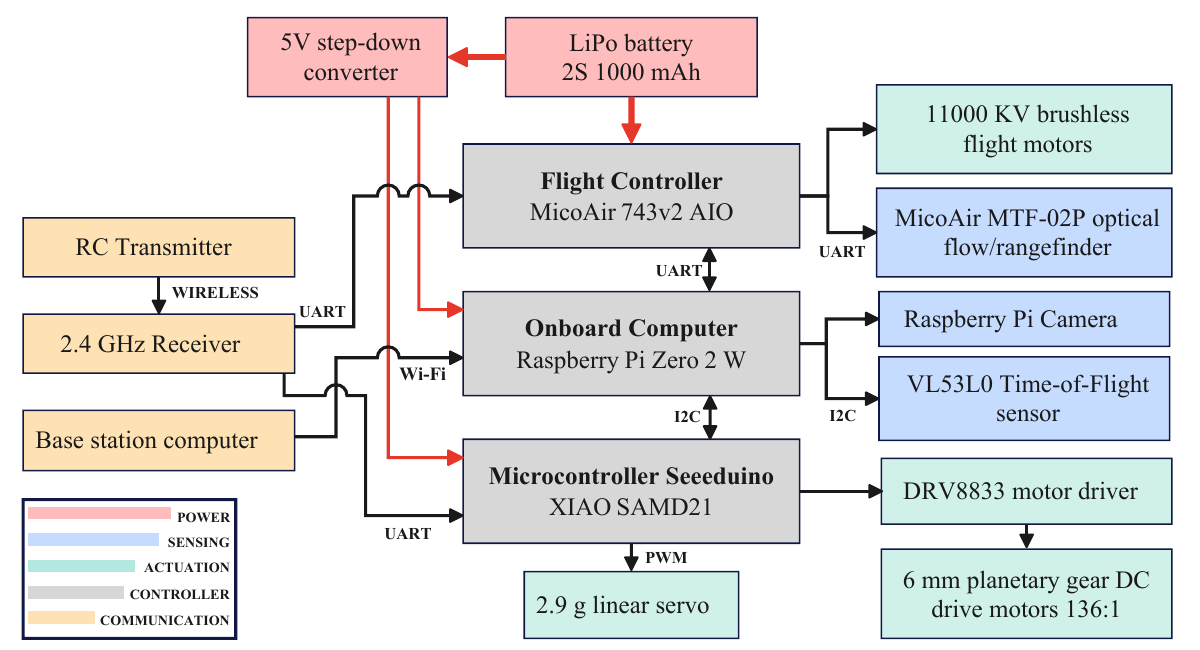}
    \caption{Electrical Architecture }
    \label{fig:electrical}
\end{figure}

AeRove uses a distributed architecture across three boards: a flight controller, a computer, and a custom printed circuit board as shown in Fig. \ref{fig:electrical}. The flight controller manages aerial actuation and onboard state estimation using the IMU, barometer, optical-flow sensor, and rangefinder. A MicoAir MTF-02P connects to the flight controller via UART and provides optical-flow and range data used for stable flight.

The custom PCB is implemented as a Raspberry Pi HAT and provides a compact interface between the onboard computer and the lower-level ground systems. It houses a Seeeduino XIAO microcontroller and a DRV8833 motor driver, which together control the terrestrial drivetrain and transition mechanism. The HAT connects directly to the Raspberry Pi through the GPIO header, with I2C used for communication between the Pi and the onboard microcontroller.

A Raspberry Pi Zero 2 W is the central onboard computing hub, routing commands to each subsystem for low-level control. It interfaces with the forward-facing Time-of-Flight (ToF) sensor through I\textsuperscript{2}C and with the Raspberry Pi camera through the ribbon-cable interface. The ToF sensor provides short-range obstacle measurements, while the camera supplies visual data for navigation and inspection. An SCD-41 CO\textsubscript{2} sensor was used for gas-sensing experiments. %

A \qty{7.4}{\volt}, 2S \qty{1000}{mAh} battery directly powers the flight controller, and is stepped down to \qty{5}{\volt} to power the Raspberry Pi, custom PCB, sensors, and auxiliary actuators. %
Manual control and mode selection are provided through a Happymodel \qty{2.4}{GHz} ELRS receiver. The receiver output is connected to both the flight controller and the microcontroller, allowing a single handheld transmitter to command either aerial or terrestrial operation. Higher-level perception and autonomy are handled through a Wi-Fi-connected base station, with the communication and control architecture described in the following subsection.

\subsection{Control and Autonomy}

AeRove uses a control architecture that separates high-level perception and mission planning from low-level vehicle actuation. The Raspberry Pi communicates with the flight controller through MAVLink for aerial commands and telemetry, while I$^2$C is used to send drivetrain and transition commands to the ground-control microcontroller. The onboard camera streams video over Wi-Fi to the base station using RTSP, where perception processing and mission-state decisions are performed. The resulting high-level commands are returned to the Raspberry Pi through a TCP connection and routed to the appropriate subsystem. Manual control and mode override are provided through the ELRS receiver.

During terrestrial locomotion, the forward-facing camera estimates the lateral position of the pipe relative to the vehicle. A PD controller converts this error into differential wheel commands, with a deadband used to suppress unnecessary correction near the pipe centerline. The camera image is divided conceptually into near- and far-field regions: the near field is used for steering correction, while the far field is used to identify larger changes in pipeline geometry such as bends, terminations, or upcoming vertical sections.

In flight, ArduPilot performs low-level attitude and thrust stabilization, while the autonomy layer issues higher-level velocity and mode commands. The forward Time-of-Flight sensor is used to identify nearby obstacles and discontinuities, while the optical-flow/rangefinding module provides local motion and altitude information. The autonomy layer therefore determines when to drive, transition, climb, cross an obstacle, descend, or resume terrestrial locomotion.

Navigation is coordinated through a hierarchical state machine that combines camera, IMU, ToF, and range measurements. When a non-traversable obstacle is detected, AeRove transitions to flight and moves forward until the ToF sensor indicates that the obstruction has been cleared. The vehicle then descends, lands on the opposite side, and resumes ground inspection. Pipe turns are handled using camera perception together with relative IMU yaw. At the start of a turn, the current yaw is stored as a local reference, and the vehicle pivots until the commanded angular change is reached. Camera-based correction is then used to recenter the platform on the pipe. Because AeRove does not rely on GPS or another absolute position reference, navigation is based on local environmental events and relative state estimates rather than continuous global-position tracking.

\begin{figure*}[!t]
    \centering
    \includegraphics[width=0.85\linewidth, clip]{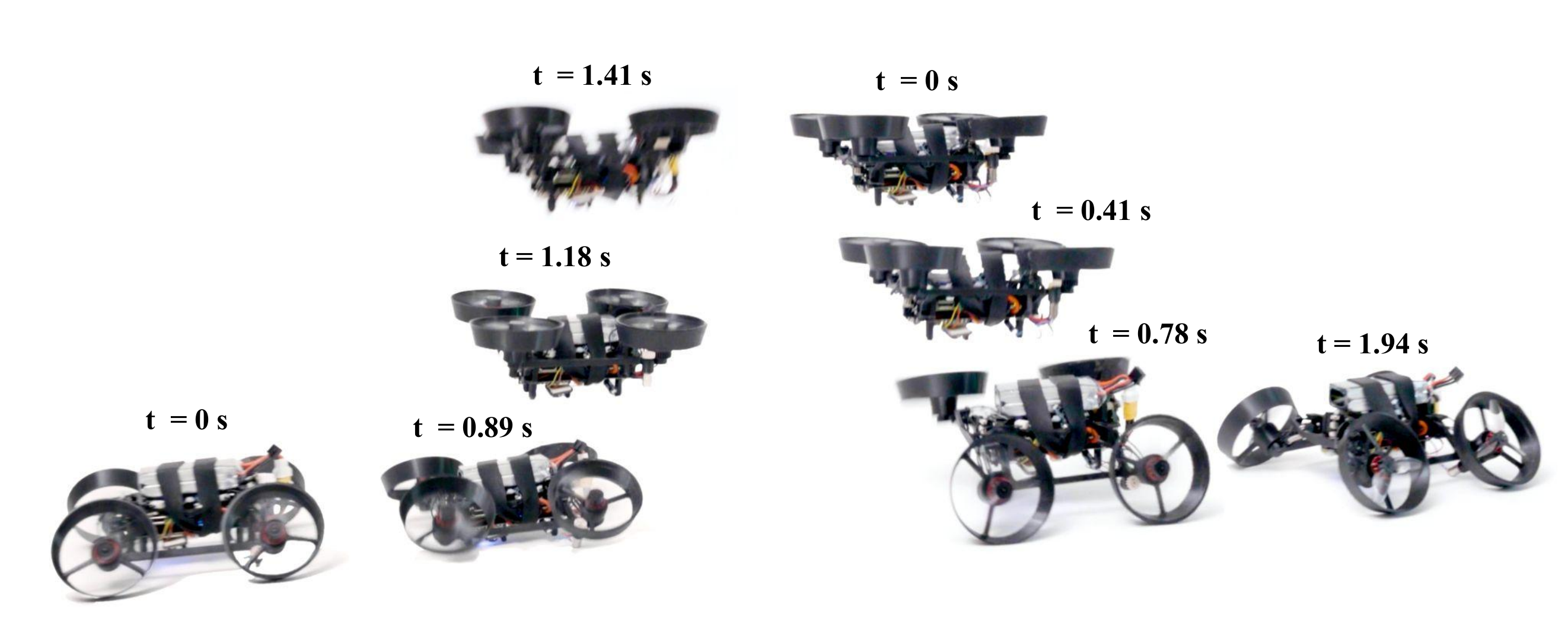}
    \caption{(Left) AeRove transitioning from terrestrial to flight mode and taking off. (Right) AeRove transitioning from flight to terrestrial mode before driving. Note that the actual transition between modes is \qty{200}{\milli\second}.}
    \label{fig:actiontrans}
\end{figure*}

\section{Experimental Testing}

The experimental evaluation was designed to answer three questions central to AeRove’s intended use: (1) whether terrestrial locomotion provides a meaningful endurance advantage over continuous flight, (2) whether the complete platform can autonomously coordinate driving, transition, and flight during representative pipeline inspection tasks, and (3) how the vehicle’s operating mode and propeller-induced airflow affect gas-sensing performance. The following experiments therefore characterize baseline locomotion performance, autonomous pipeline navigation, and gas detection under both terrestrial and aerial operating conditions.

\subsection{Basic Capability Characterization}

AeRove was characterized to quantify its baseline physical and operational capabilities in both terrestrial and aerial locomotion modes. Table \ref{tab:physical_specs} summarizes the platform's specifications, while Table \ref{tab:mode_performance} compares the two locomotion modes' characteristics during inspection and at their limits. Table \ref{tab:platform_comparison} places these values in the context of existing aerial-terrestrial platforms and representative aerial inspection vehicles.

\begin{table}[H]
\centering
\caption{AeRove Physical Specifications}
\label{tab:physical_specs}
\begin{tabular}{lc}
\hline
\textbf{Parameter} & \textbf{Value} \\
\hline
Dry Weight & 130 g \\
Battery Mass & 45 g \\
Total Mass & 175 g \\
Transition Time & 200 ms \\
Transition Angle & 75$^\circ$ \\
Thrust-to-Weight Ratio & 2.7:1 \\
\hline
\multicolumn{2}{c}{\textbf{Dimensions}} \\
\hline
Ground Mode & $12 \times 14.5 \times 5$ cm \\
Flight Mode & $14.5 \times 15.5 \times 5$ cm \\
\hline
\end{tabular}
\end{table}

\begin{table}[H]
\centering
\caption{Estimated Operating Performance}
\label{tab:mode_performance}
\setlength{\tabcolsep}{3.5pt}
\renewcommand{\arraystretch}{1.05}
\begin{tabular}{lcccc}
\hline
\textbf{Mode} &
\textbf{Speed} &
\textbf{Current} &
\textbf{Time} &
\textbf{Range} \\
\hline
Inspection (Drive) & 40 cm/s & 0.7 A & 85.7 min & 2057 m \\
Inspection (Flight) & 40 cm/s & 10 A & 6.0 min & 144 m \\
Max Speed (Drive) & 40 cm/s & 0.7 A & 85.7 min & 2057 m \\
Max Speed (Flight) & 8 m/s & 20 A & 3.0 min & 1440 m \\
\hline
\end{tabular}
\end{table}

\begin{table*}[t]
\centering
\caption{Comparison of AeRove with aerial-terrestrial, pure flight, and inspection platforms.}
\label{tab:platform_comparison}
\setlength{\tabcolsep}{7pt}
\renewcommand{\arraystretch}{1.15}

\begin{tabular}{lccccc}
\hline
\textbf{Platform} &
\textbf{Weight} &
\textbf{Flight Endurance} &
\textbf{Drive Endurance} &
\textbf{Flight Speed} &
\textbf{Ground Speed} \\
\hline

AeRove &
175 g &
6 min &
84 min &
9 m/s &
0.4 m/s \\

AeRove (flight hardware only) &
130 g &
12.5 min &
N/A &
\textgreater9 m/s &
N/A \\

Micro Aerial-Ground Robot~\cite{araki2017multirobot} &
41 g &
5 min &
42 min &
0.3 m/s &
0.1 m/s \\

Drivocopter~\cite{kalantari2020drivocopter} &
4500 g &
8 min &
90 min &
-- &
1.0 m/s \\

HyTAQ~\cite{kalantari2013hytaq} &
450 g &
7 min &
27 min &
2.0 m/s &
1.5 m/s \\

Comparison quadrotor~\cite{flywoo_flylens85} &
69 g &
9 min &
N/A &
21 m/s &
N/A \\

Voliro~\cite{voliro2026} &
6000 g &
10--14 min &
N/A &
15 m/s &
N/A \\

\hline
\end{tabular}
\end{table*}

To estimate mission range as the proportion of aerial travel varies, battery-capacity consumption was modeled as

\begin{equation}
C(d)
=
d\left[
(1-f)\frac{I_d}{3600\,v}
+
f\frac{I_f}{3600\,v}
\right]
\label{eq:current_consumption}
\end{equation}

where \(d\) is the total distance traveled, \(f\) is the fraction of the mission completed in flight, \(I_d\) and \(I_f\) are the measured current draws in drive and flight modes, respectively, and \(v\) is the inspection speed.

\subsection{Pipeline Testing Results}

To evaluate AeRove in a representative pipeline environment, an indoor testbed was constructed using a pipe with diameter \qty{51}{\centi\meter} (\qty{20}{in}) with a \qty{2.4}{\meter} (\qty{8}{ft}) horizontal section and a \qty{1.8}{\meter} (\qty{6}{ft}) vertical section. An obstacle was placed along the horizontal pipe to represent a pipeline obstruction such as a flange or valve. For safety reasons, instead of releasing propane or natural gas for detection, we used green LEDs to represent simulated gas-leak indicators, and red LEDs to represent camera-detectable surface damage.
The testbed was designed to reproduce several conditions expected in industrial pipeline networks, including horizontal traversal, directional changes, vertical sections, and localized obstructions.

Three autonomous inspection trials were conducted. The first evaluated ground-only traversal through a straight section and a 90$^\circ$ right-hand turn. The second evaluated traversal of a straight horizontal section followed by detection of a vertical pipe segment and transition to flight for continued inspection. The third focused on obstacle negotiation, requiring AeRove to detect the obstruction, transition from ground locomotion to flight, cross the obstacle, land on the opposite side, and resume terrestrial inspection.

\begin{figure}
    \centering
    \includegraphics[width=\linewidth, trim=0 4cm 0 0cm, clip]{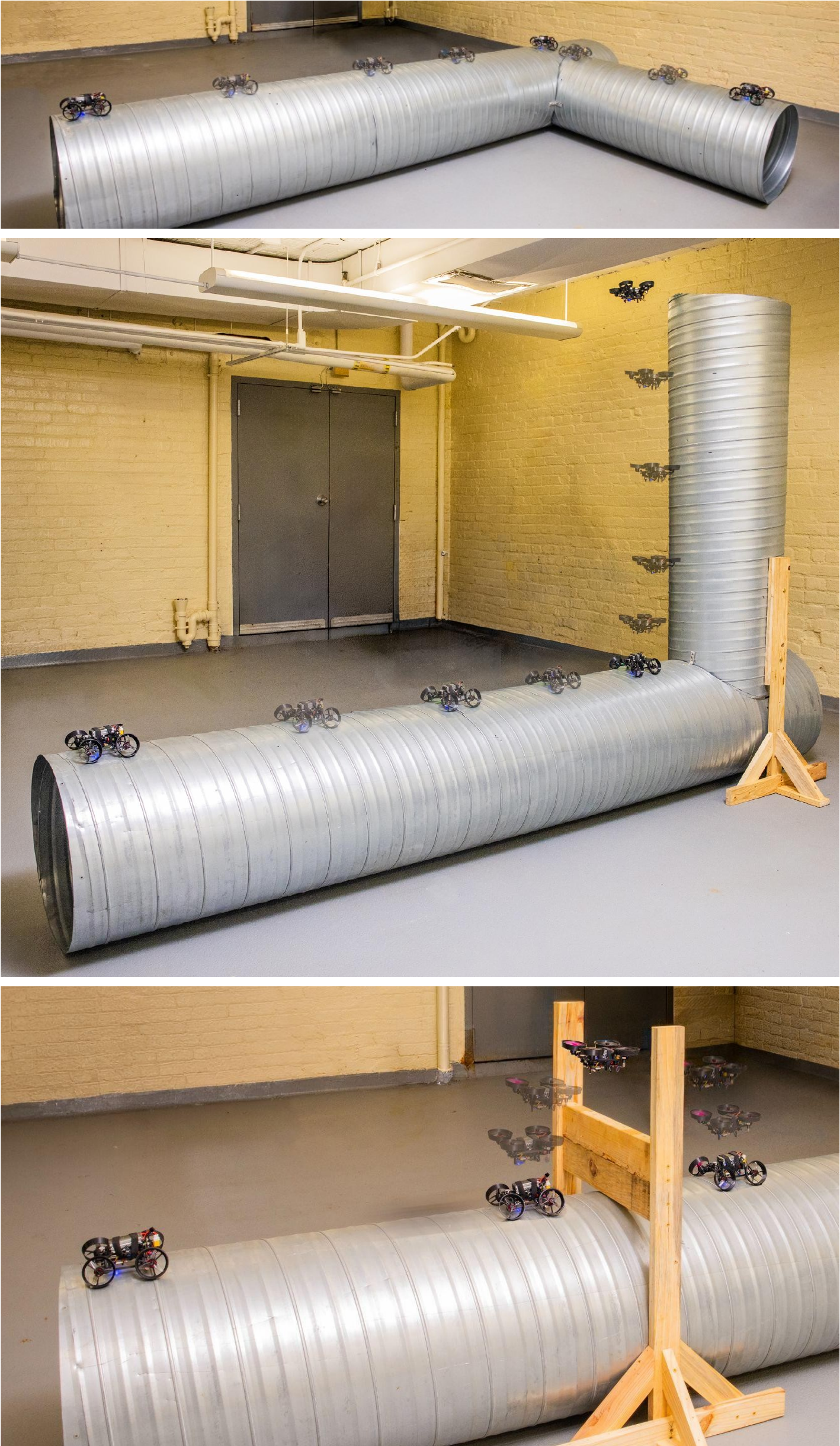}
    \caption{Autonomous pipeline inspection trials: (top) terrestrial traversal through a 90° bend; (middle) transition from horizontal ground inspection to aerial inspection of a vertical pipe section; (bottom) obstacle detection, aerial traversal, landing, and resumed terrestrial locomotion.}
    \label{fig:pipeline}
\end{figure}

Across the three trials, AeRove demonstrated autonomous straight-line driving, pipeline turning, detection of vertical sections, aerial inspection of vertical pipe geometry, obstacle detection and clearance, and return to ground locomotion after flight shown in Fig. \ref{fig:pipeline}. The platform also correctly identified the simulated gas-leak and damage indicators and reported the corresponding events to the base station.

\begin{figure}
    \centering
    \includegraphics[width=\linewidth]{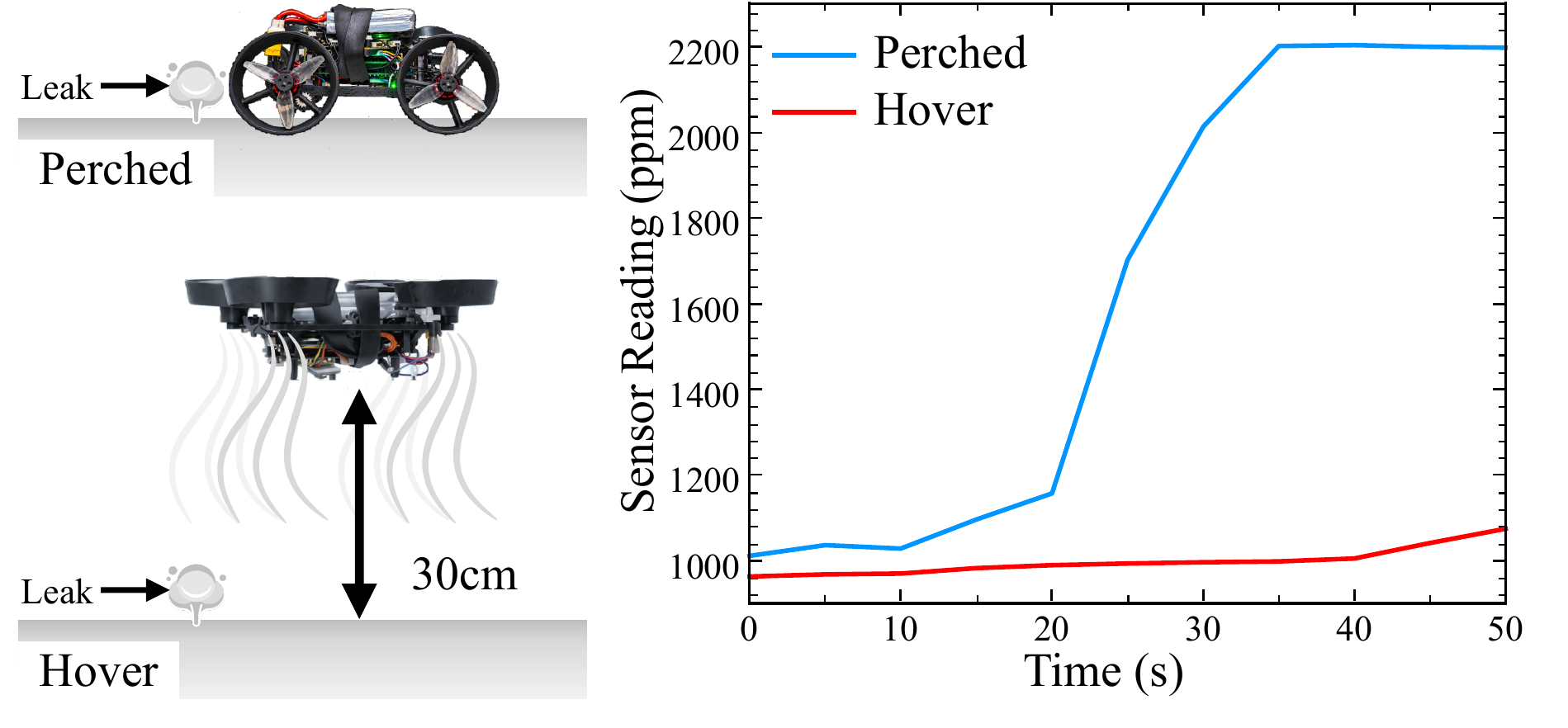}
    \caption{Representative data for perched and hovering measurements. Graph shows CO\textsubscript{2} detection in parts per million.}
    \label{fig:gastesting_h}
\end{figure}

\begin{figure}
    \centering
    \includegraphics[width=\linewidth]{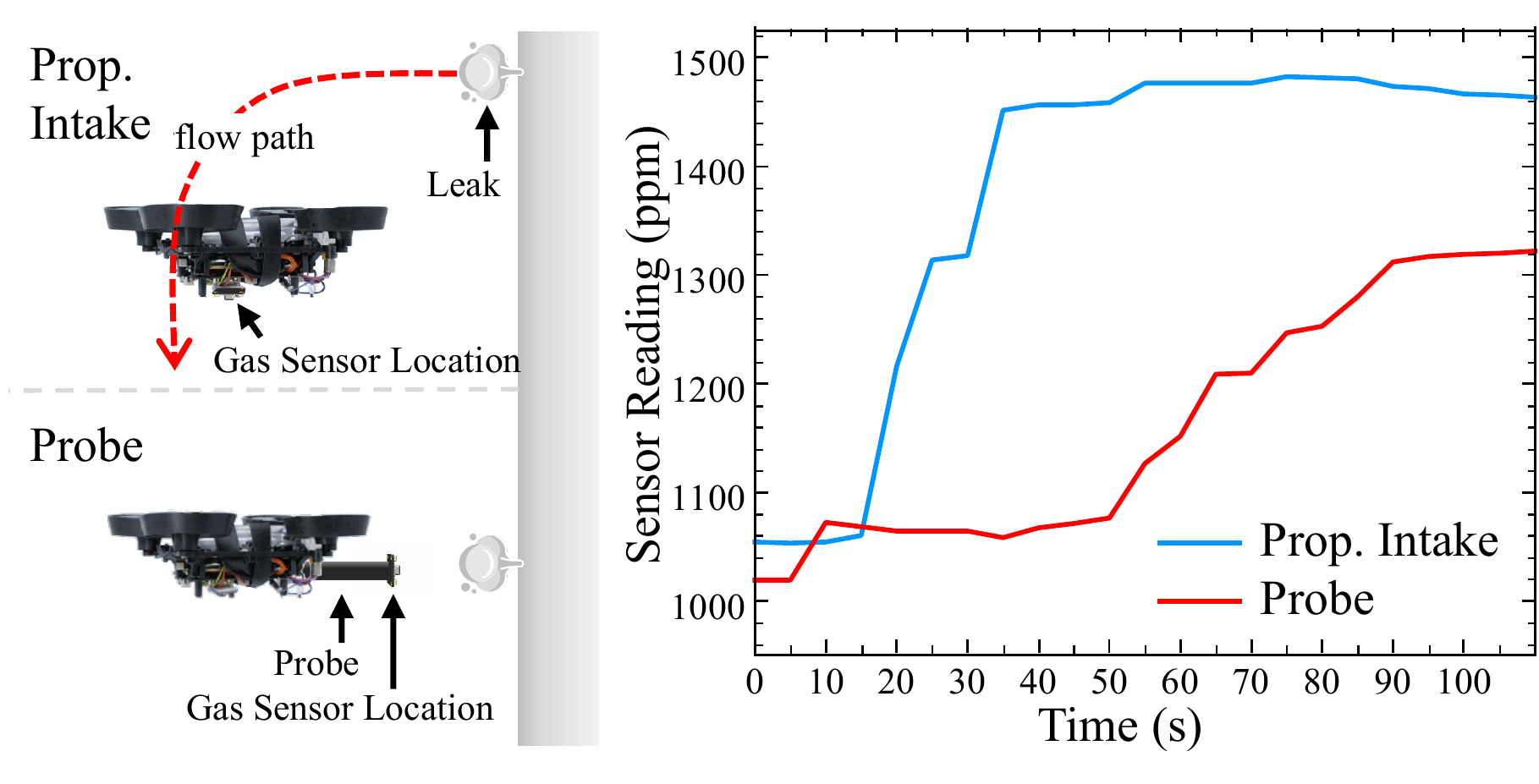}
    \caption{Representative test data for propeller intake and probe measurements. Graph shows CO\textsubscript{2} detection.}
    \label{fig:gastesting_v}
\end{figure}

\subsection{Effect of Propeller Flow on Gas Sensing}

Prop wash can strongly interfere with gas sensing by dispersing the local concentration field around the vehicle, reducing the trace of gas reaching the sensor. 

To evaluate this effect and compare sensing strategies, carbon dioxide was used as a safe surrogate for propane during indoor testing. The onboard SCD-41 CO$_2$ sensor was mounted beneath the vehicle. Although CO$_2$ is not chemically representative of propane, its comparable molecular mass provides more similar buoyancy behavior than methane while offering a safe and repeatable surrogate for relative sensing comparisons under controlled indoor conditions without flammability concerns. Validation with representative propane or natural-gas dispersion will therefore be required in future work.

The first experiment compared gas detection while AeRove was perched on a horizontal pipe with detection while hovering \qty{30}{\centi\meter} above the same location. The ambient CO$_2$ concentration was approximately \qty{1000}{ppm}, and CO$_2$ was released from a canister for two seconds at the beginning of each trial. As shown in Fig.~\ref{fig:gastesting_h}, the hovering configuration produced only a small increase in measured concentration, reaching a peak of approximately \qty{1071}{ppm}. In contrast, the perched configuration reached approximately \qty{2200}{ppm} under the same test conditions and responded more rapidly to the release. This difference reflects the combined effects of the operating configuration, including both reduced sensor-to-source distance and the absence of continuous rotor-induced airflow.

A second experiment evaluated gas sensing during vertical inspection, where continuous flight is required. Two sensor configurations were compared. In the first, the sensor remained beneath the vehicle so that air drawn into the propellers from above could be directed toward the detector. This configuration was motivated by the inspection sequence: as AeRove ascends, newly inspected pipe surface lies above the vehicle, allowing gas released from an uninspected region to enter the rotor inflow. The second configuration placed the sensor on a \qty{50}{\centi\meter} probe extending away from the propeller wash. CO$_2$ was released for five seconds while the vehicle hovered, and the sensor response was recorded for each configuration.

As shown in Fig.~\ref{fig:gastesting_v}, the propeller-intake configuration produced a faster response and a higher measured concentration than the extended probe under the tested conditions. The probe response increased more gradually, but it remained physically separated from the strongest propeller-induced flow. These results suggest that sensor placement and vehicle aerodynamics can be used strategically during aerial inspection, while terrestrial or perched operation remains preferable when the goal is to detect small or localized gas leaks.

\section{Discussion}

AeRove's primary advantage is not simply its ability to both drive and fly, but its ability to use each locomotion mode where it provides the greatest benefit. At the same nominal inspection speed of 0.40 m/s, ground locomotion draws approximately 0.7 A, compared with 10 A in flight, corresponding to a 14.3$\times$ reduction in current draw. AeRove can therefore remain in terrestrial mode during routine traversal and reserve flight for obstacles, discontinuities, or sections that cannot be negotiated from the surface.

The bistable transition mechanism supports this operating strategy by enabling rapid reconfiguration without continuous actuator power to hold either state. During the flight-to-ground transition, reconfiguration can begin before touchdown so that the wheel configuration is established prior to surface contact, reducing the chance that uneven terrain interferes with the mechanism. The short transition time also minimizes the interruption between aerial and terrestrial portions of the inspection sequence.

Hybridization introduces additional mass from the terrestrial drivetrain and transition hardware that would not be present on a purely aerial platform. This tradeoff is evident when comparing the complete AeRove platform with its flight-only configuration: removing the terrestrial hardware reduces the vehicle mass from 175 g to 130 g and increases estimated flight endurance from 6 min to 12.5 min. However, the complete platform provides a substantially longer-duration terrestrial mode, allowing flight to be reserved for obstacles and discontinuities rather than continuous traversal. As shown in Table~\ref{tab:platform_comparison}, this operating strategy is consistent with the endurance advantage observed in other aerial-terrestrial platforms, where ground locomotion can provide considerably longer operating times than flight. AeRove therefore accepts reduced aerial endurance in exchange for a low-power ground mode that can extend mission-level operation when most of the inspection route is terrestrially traversable. The endurance values reported for AeRove are estimated from the average measured current draw of each operating mode rather than from complete battery-discharge or autonomous mission trials and should therefore not be interpreted as measured full-mission endurance.

The gas-sensing experiments also demonstrate how locomotion mode can influence sensing performance. When surface traversal is possible, perched operation avoids continuous propeller-induced airflow and produced a substantially larger concentration response than hovering under the tested conditions. When flight is required, however, the propeller flow can be used as part of the sensing strategy rather than treated only as a disturbance. The underbody intake configuration produced a faster and stronger response than the extended probe, indicating that sensor placement should be designed in conjunction with the vehicle's aerodynamic flow field. These results support terrestrial inspection when localized gas sensitivity is important, while also providing a practical sensing approach for unavoidable aerial segments.

Although gas-pipeline inspection was used as the primary application in this work, the same locomotion architecture could be applied to other confined or hazardous inspection environments. Potential applications include bridges, tunnels, ventilation systems, and other infrastructure where extended surface traversal is possible but occasional aerial motion is required to cross discontinuities or avoid obstacles.

Further validation is required under more representative pipeline conditions. Future testing should evaluate the complete autonomous system across different obstacle geometries, pipe bends, pipe diameters, lighting conditions, surface irregularities, and external disturbances. Testing across multiple pipe diameters is particularly important because the current terrestrial platform was originally developed for more general surface locomotion and was not optimized specifically for cylindrical geometry. Driving performance on the curved pipe surface was therefore more difficult than on planar terrain. Future mechanical iterations should optimize wheel placement, contact geometry, and traction for cylindrical surfaces while maintaining the transition and flight capabilities demonstrated in this work.

\section{Conclusion}

This paper demonstrated AeRove, a bimodal ground-flight platform over %
three autonomous inspection scenarios that included terrestrial traversal around a \qty{90}{\degree} corner; mixed-mode traversal over an obstacle; and vertical traversal near a vertical pipe. Additional experiments demonstrated detection of simulated leak and damage markers, while separate CO\textsubscript{2} experiments evaluated how operating configuration and propeller-induced airflow affect gas sensing both positively and negatively.
The results show that combining low-power terrestrial locomotion with selective aerial mobility can extend mission range while retaining the ability to traverse discontinuities that cannot be crossed with wheeled motion. The bistable transition mechanism supports this approach by enabling rapid reconfiguration without continuous actuator power in either locomotion state. %

Future work will focus on improving both mobility and autonomy. Planned developments include attachment mechanisms for reliable operation on walls, ceilings, and other surfaces, cooperative behaviors that allow multiple AeRove units to coordinate inspection tasks, and a more capable autonomy framework for navigating complex pipe networks and a wider range of operating conditions.

\section*{Acknowledgment}
We would like to thank Dr. Pramod Abichandani, Prof. Christian Hansis, and Dr. Raphael Zufferey for their technical assistance throughout the project.
This work was supported by Nation Science Foundation Award \#2452560 as well as by the NJIT Undergraduate Research and innovation program.
We would like to specifically thank Orlando Castillo for his assistance in creating the test environment, AEROJO Drone Productions for his assistance in photography, and GrayBeard Solutions LLC for the 3D printing services. \textit{AI Disclosure}: Gemini 3.6 Thinking was used to help condense several paragraphs during final editing. %


\end{document}